\documentclass[pdflatex,sn-basic]{sn-jnl}
\usepackage{amsmath,amssymb}%
\usepackage{booktabs,tabularx}%
\usepackage{xcolor}%
\let\orcidlogo\relax
\usepackage{orcidlink}%
\usepackage{tikz}%
\usetikzlibrary{positioning,arrows.meta,fit,backgrounds,calc}%
\usepackage{pgfplots}%
\pgfplotsset{compat=1.18}%

\definecolor{oldred}{RGB}{176,58,46}
\definecolor{newgreen}{RGB}{30,123,90}
\definecolor{libblue}{RGB}{52,92,150}
\newsavebox{\codebox}
\begin{document}

\title{Unmasking Four Knapsack Problems in the Traveling Thief Benchmark}

\author[1]{\fnm{Mohammed Mizan} \sur{Alanazi}}

\author*[1,2,3]{\fnm{Mahdi} \sur{Khemakhem}\,\orcidlink{0000-0001-5603-1947}}\email{mahdi.khemakhem@enetcom.usf.tn}

\affil*[1]{\orgdiv{Department of Computer Science, College of Computer Engineering and Sciences}, \orgname{Prince Sattam
bin Abdulaziz University}, \orgaddress{\city{AlKharj}, \postcode{11942}, \country{Saudi Arabia}}}

\affil[2]{\orgname{National School of Electronics and Telecommunications of Sfax (ENET'Com), University of Sfax}, \orgaddress{\city{Sfax}, \country{Tunisia}}}

\affil[3]{\orgdiv{Data Engineering and Semantics Research Unit, Faculty of Sciences of Sfax}, \orgname{University of Sfax}, \orgaddress{\city{Sfax}, \country{Tunisia}}}

\abstract{The traveling thief problem (TTP) asks a thief to choose a tour and a packing together, and the benchmark of
9,720 instances of Polyakovskiy et al.\ (2014) is the standard test set for comparing TTP heuristics. We show that four
of these instances are not TTP instances at all. Their files give a renting rate of 0.00, so the tour has no effect on
the objective and each instance reduces to a 0-1 knapsack problem. One of them belongs to the 60 categorized instances
on which new methods are usually compared, and on it five of nine published methods report exactly the same value.
The cause is the format of the files, which keep the renting rate with two decimals, while the rule of the generator
gives a rate close to 0.003 on these instances. The same rounding may shift the rate of 949 other instances by more
than 1\%. Since the tour and the packing used by the generator were not published, the original rates cannot be
recovered. We therefore define new rates with the same rule, applied to a fully specified reference solution that we
publish, and we show with a published heuristic that the corrected instances behave as TTP instances again. We also
give reference results on the corrected instances and recommendations for future studies. Finally, we present
\texttt{ttplib}, a new open-source Python library that loads any of the 9,720 instances, including the corrected ones,
without downloading the whole benchmark.}


\maketitle

\section{Introduction}
\label{sec:intro}
In the traveling thief problem \citep{bonyadi2013travelling}, a thief visits every city once and fills a knapsack with
items found along the way. We use the formulation of the benchmark \citep{polyakovskiy2014comprehensive}. An instance
has $n$ cities, numbered $0,\dots,n-1$ with the depot at city~$0$, and $m$ items. Item $j$ has a profit $p_j$ and a
weight $w_j$, and it lies in city $a_j\neq 0$. The knapsack holds at most $W$, the rent is $R$ per unit of time, and the
speed of the thief falls linearly with the carried weight $c$,
\begin{equation}
v(c)=v_{\max}-\nu c,\qquad \nu=\frac{v_{\max}-v_{\min}}{W},
\label{eq:speed}
\end{equation}
from $v_{\max}$ when empty to $v_{\min}$ when full. A solution is a tour $\pi$ that visits every city once and returns to the depot, written as the sequence of cities
$\pi_0,\pi_1,\dots,\pi_n$ with $\pi_0=\pi_n=0$, together with a set $S$ of picked items with
$\sum_{j\in S}w_j\le W$. Edge $s$ of the tour goes from $\pi_s$ to $\pi_{s+1}$, for $s=0,\dots,n-1$. We note $d_s$ for
its length, which is the Euclidean distance rounded up to the next integer, and $c_s$ for the weight carried on it,
which is the total weight of the items picked in the cities $\pi_0,\dots,\pi_s$. The objective, to be maximized, is
the profit minus the rent paid for the travel time,
\begin{equation}
Z(\pi,S)=P(S)-R\,T(\pi,S),\qquad P(S)=\sum_{j\in S}p_j,\qquad T(\pi,S)=\sum_{s=0}^{n-1}\frac{d_s}{v(c_s)}.
\label{eq:ttp}
\end{equation}

The benchmark of \citet{polyakovskiy2014comprehensive}\footnote{The original files of the 9,720 instances are available
at \url{https://cs.adelaide.edu.au/~optlog/CEC2014COMP_InstancesNew/}, accessed on 27 October 2026.} crosses 81
instances of the TSP library with four numbers of items per city, three types of knapsack and ten capacity levels, which
gives 9,720 instances. The capacity level, from 01 to 10, is the last part of the name of an instance. It was used in the IEEE CEC 2014 competition \citep{polyakovskiymanual}. Since then, few studies
have used the full set, and different studies have chosen different subsets \citep{wagner2018case}. The full set is
too large for the usual protocol of ten runs of ten minutes per instance, which would take about 1.8 years of
computing time for a single method. Most recent comparisons therefore use 60 \emph{categorized instances}
sampled from the 9,720, a sample introduced by \citet{el2016population} following the sample instances of the CEC 2014
competition and used,
for example, by \citet{el2018efficiently} and
\citet{namazi2019cooperative,namazi2023solving}. This sample takes twenty TSP instances, from eil76 with 76 cities to
pla33810 with 33,810 cities, so that it covers small, medium and large instances. On each of them, it takes one TTP
instance of each of three categories, which represent three different kinds of difficulty. Category~A has one item
per city, a bounded strongly correlated knapsack and the smallest capacity (level~01), so few items can be picked and
the tour has a large weight in the objective. Category~B has five items per city, an uncorrelated knapsack with
similar weights and a medium capacity (level~05). Category~C has ten items per city, an uncorrelated knapsack and the
largest capacity (level~10), so the packing has the most choices. Published values exist for many methods on these 60
instances, which is the main reason why new methods are compared on them. \citet{sarkar2024travelling} survey the
variants of the problem and the methods proposed for it.

The rent $R$ links the two parts of the problem. If $R$ is large, the travel time dominates the objective. If $R$ is
small, the profit dominates. If $R=0$, the second term of Eq.~(\ref{eq:ttp}) vanishes and $Z(\pi,S)=P(S)$ for every tour
$\pi$. Maximizing $P(S)$ under the capacity constraint is the 0-1 knapsack problem. Any tour can be used, and the best
solution is an optimal packing of the knapsack part \citep{wagner2018case}. Some tables of instances in the literature
list a rent of 0.00, for example for two of the dsj1000 instances in \citet{ali2021novel}, and
\citet{nikfarjam2024use} left aside the first dsj1000 instances because their rent is zero, which turns them into
knapsack problems. To the best of our knowledge, no study has explained where these zeros come from, listed all the
instances concerned, examined the effect of the zero rent on published results, or proposed a correction.

We discovered this case while working on a new method to solve the TTP. On the instance of category~A built on dsj1000,
repeated runs returned exactly the same value, 144,426, which is the optimal profit of the knapsack part, and the file
of this instance gives $R=0.00$. We also noticed that, on this instance, most of the methods compared in recent studies
report values very close to this optimum, and five of them report exactly the same value. We then read all 9,720 files
to find every instance with this problem, and to measure how far the way the rent is stored affects the others.

This note aims to make the problem known to the community and to give a simple way to fix it.
Section~\ref{sec:zero} describes the four instances and explains why their rent is zero. Section~\ref{sec:rates}
defines new rents and gives reference results on the corrected instances. Section~\ref{sec:library} presents the
corrected files and \texttt{ttplib}, a new library that we developed, and Section~\ref{sec:recommend} concludes with recommendations.

\section{Four instances with a zero rent}
\label{sec:zero}

\subsection{The four instances and their published results}
\label{sec:four}
We read the line \texttt{RENTING RATIO} of all 9,720 \texttt{.ttp} files. In every file, the value is written with exactly
two decimals. Four files give $R=0.00$ (Table~\ref{tab:zero}). All four have 1,000 cities and 999 items, and the last two columns
of Table~\ref{tab:zero} give the profit and the number of items of the optimal packing of the knapsack part. With $R=0$, the objective of any solution is the total
profit of the picked items, whatever the tour. Each of the four instances is then a 0-1 knapsack problem, not a TTP
instance.

\begin{table}[ht]
\centering\footnotesize
\caption{The four instances with $R=0.00$.}
\label{tab:zero}
\begin{tabularx}{\linewidth}{@{}Xrrr@{}}
\toprule
\textbf{Instance} & \textbf{Capacity} & \textbf{Optimal profit} & \textbf{Items packed} \\
\midrule
dsj1000\_n999\_bounded-strongly-corr\_01 & 87,826 & 144,426 & 291 \\
dsj1000\_n999\_uncorr-similar-weights\_01 & 91,232 & 85,721 & 90 \\
dsj1000\_n999\_uncorr-similar-weights\_02 & 182,464 & 164,663 & 181 \\
dsj1000\_n999\_uncorr\_01 & 45,302 & 165,552 & 257 \\
\bottomrule
\end{tabularx}
\end{table}

The first of the four is one of the 60 categorized instances, the instance of category~A built on dsj1000. Every
method evaluated on the categorized instances has solved a pure knapsack problem on that instance. Its best possible value is the
optimal knapsack profit, 144,426, which we computed exactly by dynamic programming and checked with Gurobi.

This instance stands out in the published results. Table~\ref{tab:published} gives the values reported for it by nine
methods that are often compared on the categorized instances. The value of CoCo$^+$ is the best value over all runs of the CoCo
variants, and the other values are means over ten runs. The values of MATLS, RWS and S5 are taken from Table~1 of
\citet{zhang2021solving}. Since the rent is zero, the tour and the way distances are
computed play no role, so these values can be compared directly. Five of the nine methods report exactly the same value,
144,219, which is 207 below the optimum, and only the best value of CoCo$^+$ reaches 144,426. On the other 59 instances
of the categorized instances, no value is shared by more than two of the nine methods, and on 55 of them all nine values differ. With a
zero rent, the methods differ only in the packing they find, and the instance measures a knapsack heuristic, not the
coordination of tour and packing that the TTP is meant to test. We did not find a study that points out the effect
of the zero rent of this instance on the published comparisons.

\begin{table}[ht]
\centering\footnotesize
\caption{Published values on dsj1000\_n999\_bounded-strongly-corr\_01.}
\label{tab:published}
\begin{tabularx}{\linewidth}{@{}lXr@{}}
\toprule
\textbf{Method} & \textbf{Source of the value} & \textbf{Value} \\
\midrule
CoCo$^+$ & \citet{namazi2023solving} & 144,426 \\
CoCo & \citet{namazi2019cooperative} & 144,219 \\
CS2SA & \citet{el2016population} & 144,219 \\
MA2B & \citet{el2016population} & 144,219 \\
CS2SA$^\ast$ & \citet{el2018efficiently} & 144,219 \\
SAVI & \citet{nguyen2023simulated} & 144,219 \\
MATLS & \citet{mei2014improving}, value reported by \citet{zhang2021solving} & 143,699 \\
RWS & \citet{zhang2021solving} & 140,656 \\
S5 & \citet{faulkner2015approximate}, value reported by \citet{zhang2021solving} & 137,653 \\
\bottomrule
\end{tabularx}
\end{table}

Does this instance change the published rankings? We computed the mean rank of the nine methods of
Table~\ref{tab:published} on the values reported by the same sources on the 60 categorized instances, which are best values for
CoCo$^+$ and means for the other methods, with and without this instance. Tied values share the mean of their ranks.
Over the 60 instances, the mean ranks move by at most 0.06 (for S5). The order of the methods stays the same, except
for RWS and SAVI, whose mean ranks differ by less than 0.04 and change places. Over the 20 instances of category~A, the
mean ranks move by at most 0.26 (for S5), and the order of the nine methods stays the same. In this simple descriptive
calculation, the effect of the instance is small. We did not repeat the statistical tests of the original studies.

\subsection{The zero rents in three published studies}
\label{sec:studies}
Three studies published in the \textit{Journal of Heuristics} show how the problem appears in practice.

\citet{wagner2018case} studied 21 algorithms on all 9,720 instances and used the results to build algorithm
portfolios, which choose the best algorithm for an instance from features of the instance. They describe the case
$R=0$ in theory and note that the best solution is then the best solution of the knapsack part, but they do not point
out that four instances of the benchmark are in this case. These four instances were therefore treated as TTP
instances, both to compare the 21 algorithms and to train the portfolios, although the tour plays no role in them. The
rent itself is one of the instance features. In their analysis of feature importance, it is the second most important
feature for choosing an algorithm, after the capacity of the knapsack. Yet this feature is zero for the four instances,
and for 217 others it has a single significant digit (Section~\ref{sec:rounding}). The feature that the portfolios
rely on most after the capacity is thus missing for four instances and imprecise for many more.

\citet{namazi2023solving} proposed the CoCo variants, whose main idea is to coordinate the changes of the tour and of
the packing, and compared them with other methods on the categorized instances. On the instance of category~A built on
dsj1000, this coordination cannot have any effect, since the tour does not change the value. Their Table~6 shows the
consequence. Over ten runs of each method, it gives a mean of 144,219 with a standard deviation of zero for
CS2SA$^\ast$, CoCoP and CoCoL, and a mean of 143,280 with a standard deviation of zero for MATLS. At its displayed
precision, the table thus reports no variation over ten runs for four methods, and three of them reach exactly the same
value. Their best value over all CoCo variants, 144,426 in their Table~5, is the optimal knapsack profit. For a fixed
packing, the value no longer depends on the tour, so a method that reaches the same packing in every run returns the
same value, whatever its tours. The zero rent does not force a method to find the same packing, but it removes the
effect of the tour, which is a main source of variation between runs. On this instance, the comparison therefore ranks
knapsack heuristics, not the coordination that the paper proposes. In our own experiments of
Section~\ref{sec:results} with CoCoP, the ten runs on each corrected file gave ten different
values.

\citet{chagas2021non} solve the bi-objective version of the problem, which minimizes the travel time and maximizes the
profit as two separate objectives. The rent does not appear in this version, and none of their nine instances is one of
the four. The issue reported here therefore concerns the single-objective TTP.

\subsection{Why the rent is zero}
\label{sec:cause}
\citet{polyakovskiy2014comprehensive} set the rent of each instance to
\begin{equation}
R=\frac{P(S^{\ast})}{T(\pi^{\mathrm{lk}},S^{\ast})},
\label{eq:rule}
\end{equation}
where $P(S^{\ast})$ is the profit of an optimal packing $S^{\ast}$ of the knapsack part, and the denominator is the
travel time defined in Eq.~(\ref{eq:ttp}) along a tour $\pi^{\mathrm{lk}}$ when the thief picks the items of $S^{\ast}$. This tour is
computed by \texttt{linkern}, the program of the Concorde TSP solver \citep{applegate2006traveling} that implements
the Chained Lin-Kernighan heuristic \citep{applegate2003chained}. It is a fast local search that returns a very good
TSP tour, but not necessarily an optimal one, and its result depends on a random seed. With this choice, the solution made of that tour and that packing has an
objective of exactly zero. The authors of the benchmark state that the rule guarantees at least one TTP solution with
zero objective. In the four files with $R=0$, the same packing has an objective equal to its positive profit. A solution
with an empty knapsack still has an objective of zero, but the calibration intended by the rule, in which the reference
solution with the optimal packing has an objective of zero, is lost.

\begin{figure}[ht]
\centering
\begin{tikzpicture}[
  font=\footnotesize, node distance=4mm and 2.6mm,
  box/.style={draw, rounded corners=2pt, align=center, text width=2.0cm, minimum height=1.25cm, line width=0.5pt},
  old/.style={box, draw=oldred, fill=oldred!7},
  new/.style={box, draw=newgreen, fill=newgreen!7},
  arr/.style={-{Stealth[length=2mm]}, line width=0.6pt}]
\node[old] (o1) {Optimal knapsack profit\\144{,}426};
\node[old, right=of o1] (o2) {Travel time on a \texttt{linkern} tour, about $5\times10^{7}$};
\node[old, right=of o2] (o3) {Rule~(\ref{eq:rule})\\$R=P/T$\\$\approx 0.003$};
\node[old, right=of o3] (o4) {Rent written with two decimals};
\node[old, right=of o4, fill=oldred!22, font=\footnotesize\bfseries] (o5) {$R=0.00$\\a pure knapsack problem};
\foreach \a/\b in {o1/o2,o2/o3,o3/o4,o4/o5} \draw[arr, oldred] (\a) -- (\b);
\node[new, below=9mm of o1] (n1) {Exact knapsack, fixed tie-breaking};
\node[new, right=of n1] (n2) {Fixed \texttt{linkern} tour, seed 1};
\node[new, right=of n2] (n3) {Rule~(\ref{eq:rule}) in full\\$R\approx0.00284$};
\node[new, right=of n3] (n4) {Reference solution with objective 0};
\node[new, right=of n4, fill=newgreen!22, font=\footnotesize\bfseries] (n5) {Corrected files and \texttt{ttplib}};
\foreach \a/\b in {n1/n2,n2/n3,n3/n4,n4/n5} \draw[arr, newgreen] (\a) -- (\b);
\node[anchor=south west, oldred, font=\footnotesize\itshape] at ($(o1.north west)+(0,0.5mm)$) {Generator of 2014};
\node[anchor=south west, newgreen, font=\footnotesize\itshape] at ($(n1.north west)+(0,0.5mm)$) {This note};
\end{tikzpicture}
\caption{How the rent of dsj1000\_n999\_bounded-strongly-corr\_01 becomes zero in the files (top), and how this note
defines its new rent (bottom, Section~\ref{sec:new}).}
\label{fig:flow}
\end{figure}
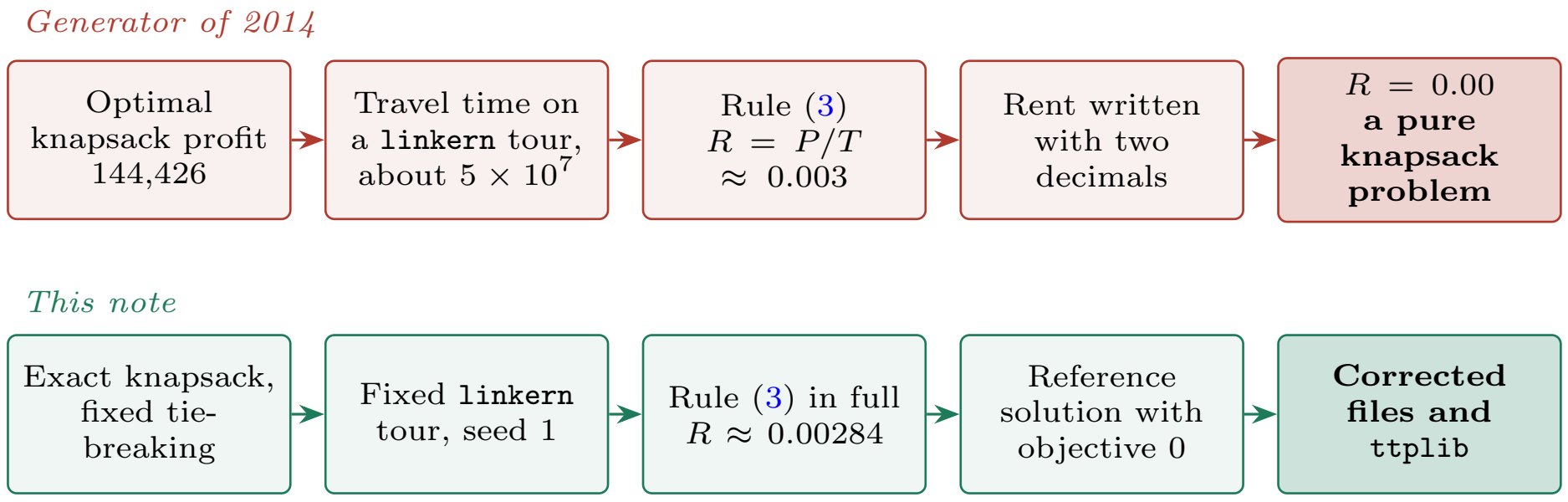

The rule itself never gives zero, since the optimal profit is positive. We cannot observe the generator, but the
available data point to the size of the ratio and to the way it was stored (Fig.~\ref{fig:flow}, top). The cities of dsj1000 spread over about $1.2\times10^6$ units in each
direction, and a good tour has a length of about $1.87\times10^7$. The thief travels at a speed between $v_{\min}=0.1$ and
$v_{\max}=1$, so the travel time is of the order of $5\times10^7$, while the optimal profit is of the order of $10^5$.
Rule~(\ref{eq:rule}) then gives a rent of about 0.003. Written with two decimals, any value below 0.005 becomes 0.00.

The files confirm this reading (Table~\ref{tab:family}). In the benchmark, the capacity of level~$k$, for
$k=1,\dots,10$, is $W=k/11$ of the total weight of the items, rounded down. For example,
dsj1000\_n999\_bounded-strongly-corr\_01 has a total weight of 966,088 and a capacity of 87,826. Table~\ref{tab:family}
gives the rent in the files of the thirty instances built on dsj1000 with 999 items, for the three types of knapsack
and the ten levels. A larger knapsack holds more profit, so the optimal profit in rule~(\ref{eq:rule}), and with it the
rent, grows with the level. The four zeros, in bold, are exactly the four instances of Table~\ref{tab:zero}: level~01
of the three types, and level~02 of the uncorrelated type with similar weights. They lie at the bottom of increasing
sequences that reach only 0.01 or 0.02 at level~10. On this TSP instance, every rent is thus at most two hundredths, so
the two decimals keep at most one significant digit, and they lose everything below 0.005.

\begin{table}[ht]
\centering\footnotesize
\caption{Rents in the files of dsj1000 with 999 items, by type of knapsack and capacity level.}
\label{tab:family}
\begin{tabular}{@{}lcccccccccc@{}}
\toprule
 & \multicolumn{10}{c}{\textbf{Capacity level}} \\
\cmidrule(l){2-11}
\textbf{Type of knapsack} & \textbf{01} & \textbf{02} & \textbf{03} & \textbf{04} & \textbf{05} & \textbf{06} & \textbf{07} & \textbf{08} & \textbf{09} & \textbf{10} \\
\midrule
bounded strongly correlated & \textbf{0.00} & 0.01 & 0.01 & 0.01 & 0.01 & 0.01 & 0.02 & 0.02 & 0.02 & 0.02 \\
uncorrelated, similar weights & \textbf{0.00} & \textbf{0.00} & 0.01 & 0.01 & 0.01 & 0.01 & 0.01 & 0.01 & 0.01 & 0.01 \\
uncorrelated & \textbf{0.00} & 0.01 & 0.01 & 0.01 & 0.01 & 0.01 & 0.01 & 0.01 & 0.01 & 0.01 \\
\bottomrule
\end{tabular}
\end{table}

\subsection{How much the rounding affects the other rents}
\label{sec:rounding}
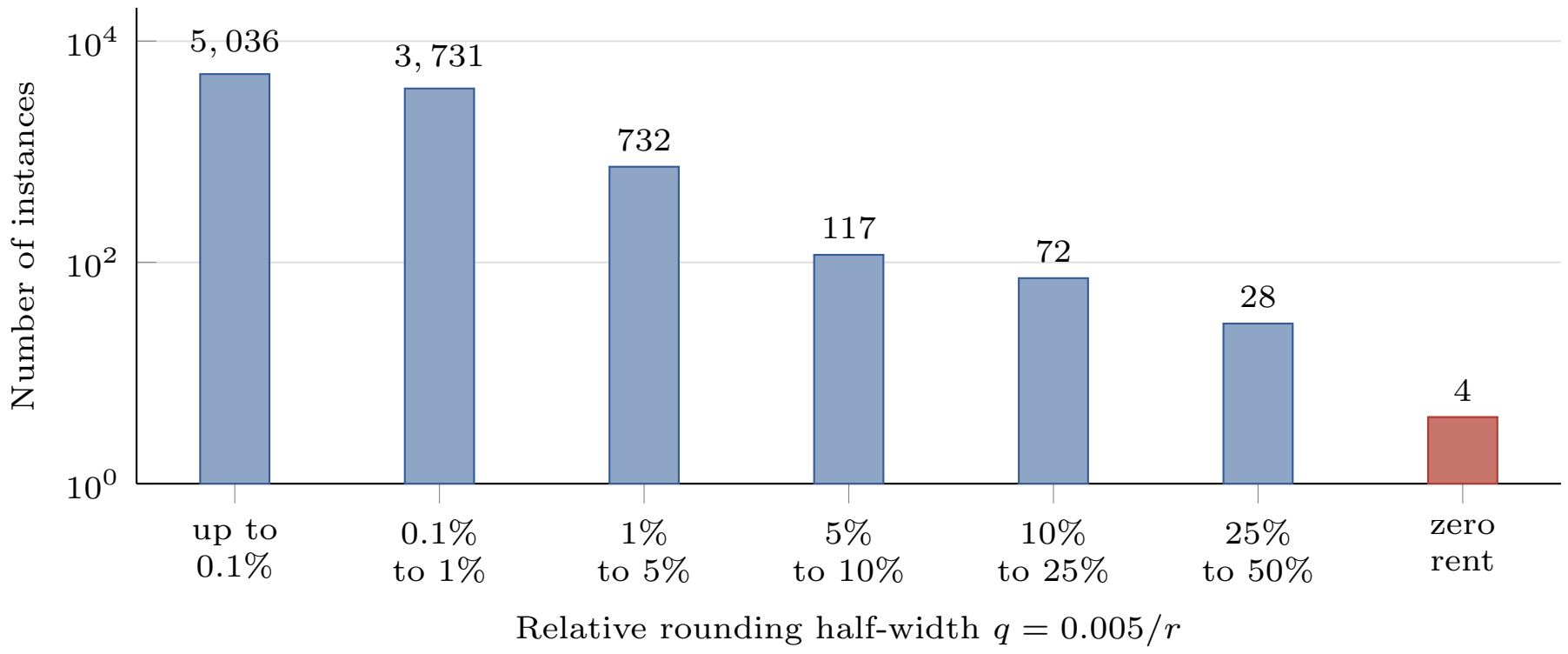
\begin{figure}[ht]
\centering
\begin{tikzpicture}
\begin{axis}[
  width=0.95\linewidth, height=5.2cm, ybar, bar width=15pt, ymode=log, ymin=1, ymax=20000,
  log origin=infty,
  symbolic x coords={a,b,c,d,e,f,g}, xtick={a,b,c,d,e,f,g}, bar shift=0pt,
  xticklabels={{up to\\0.1\%},{0.1\%\\to 1\%},{1\%\\to 5\%},{5\%\\to 10\%},{10\%\\to 25\%},{25\%\\to 50\%},{zero\\rent}},
  xticklabel style={align=center, font=\footnotesize},
  ylabel={Number of instances}, ylabel style={font=\footnotesize}, yticklabel style={font=\footnotesize},
  xlabel={Relative rounding half-width $q=0.005/r$}, xlabel style={font=\footnotesize},
  point meta=rawy, nodes near coords={\pgfmathprintnumber[fixed,precision=0,1000 sep={{,}}]{\pgfplotspointmeta}},
  nodes near coords style={font=\footnotesize},
  axis lines*=left, ymajorgrids, grid style={draw=black!12}, enlarge x limits=0.08]
\addplot[fill=libblue!55, draw=libblue] coordinates {(a,5036) (b,3731) (c,732) (d,117) (e,72) (f,28)};
\addplot[fill=oldred!70, draw=oldred] coordinates {(g,4)};
\end{axis}
\end{tikzpicture}
\caption{The 9,720 instances by relative rounding half-width $q$ (logarithmic vertical axis).}
\label{fig:rounding}
\end{figure}

Let $r$ be the rent stored in a file. If a positive rent was rounded to the nearest hundredth, its value before rounding
lies within 0.005 of $r$. For $r>0$, we call $q=0.005/r$ the relative rounding half-width. It measures how much precision
may have been lost, not the actual error, which is unknown. If the generator truncated instead, the interval has a width
of 0.01. The code of the generator is not distributed, so the calculation below assumes rounding to the nearest
hundredth. We computed $q$ for all 9,720 files (Fig.~\ref{fig:rounding}). For 52\% of the instances, $q\le0.1\%$. For
949 instances, $q>1\%$, for 217, $q>5\%$, and for 100, $q>10\%$. Twenty-eight files have $r=0.01$, giving $q=50\%$, and
another 23 have $r=0.02$, giving $q=25\%$. In Fig.~\ref{fig:rounding}, each class contains its upper end, so the 23
files with $r=0.02$ fall in the class 10\% to 25\%, and the four files with $r=0.00$ form a separate class, where $q$ is
not defined. These small rents belong to TSP instances whose tours are very long, because
their coordinates are large or their cities are many. Of the 217 instances with $q>5\%$, 98 are built on dsj1000, 36 on
pla7397, 28 on pla33810, 15 on pla85900 and 13 on usa13509.

The categorized instances are also concerned. Besides the instance with $r=0.00$, eleven of its instances have $q>1\%$, and four
of them have $q>5\%$. These four are the instances of category~A built on pla7397 ($r=0.02$, $q=25\%$), pla33810
($r=0.03$, $q\approx17\%$) and usa13509 ($r=0.04$, $q\approx13\%$), and the instance of category~B built on dsj1000
($r=0.04$, $q\approx13\%$). These instances remain proper TTP instances, so we do not change them. But the value of
their rent intended by the generator, and so the weight of the tour in their objective, may differ noticeably from the
stored value.

\section{New renting rates for the four instances}
\label{sec:rates}

\subsection{Why the original values cannot be recovered}
\label{sec:recover}
A correction should ideally restore the value that the generator computed before rounding. This value cannot be
recovered from the files and code that \citet{polyakovskiy2014comprehensive} published with the benchmark, for three
reasons. First, rule~(\ref{eq:rule}) depends on the tour, and the
authors of the benchmark distribute their tour only for the TSP instance a280. Their code does not contain the
generator, but it contains the call to \texttt{linkern} and the evaluation function, and we implemented the rule with
them, with the tour read from the output of \texttt{linkern} from the depot, the travel time computed as
in Eq.~(\ref{eq:ttp}) and the knapsack solved exactly. With their tour, the recomputed rent of the 120 instances built on
a280 has a median ratio of 0.975 to the value in the file, and 90\% of the ratios lie between 0.89 and 1.08. Rounded to
two decimals, it equals the value in the file on only one of the 120 instances, so the same implementation is close to
the files but does not reproduce them. Second, the rent changes with the tour, even among tours of almost the same
length. For each of the four instances, we ran \texttt{linkern} with ten seeds. The ten tours differ in length by less
than 1\%, but with the same optimal packing the rent varies by up to 16\% (Table~\ref{tab:spread}), because the travel
time depends on where along the tour the heavy items are picked. Third, the rent changes with the optimal packing. A
knapsack can have several optimal packings with the same profit. On the bounded strongly correlated instance, two
optimal packings give rents that differ by 2.0\% on the same tour, and the packing chosen by the original knapsack
solver \citep{martello1999dynamic} is not known.

\begin{table}[ht]
\centering\footnotesize
\caption{Rents given by rule~(\ref{eq:rule}) on ten \texttt{linkern} tours.}
\label{tab:spread}
\begin{tabularx}{\linewidth}{@{}Xcc@{}}
\toprule
\textbf{Instance} & \textbf{Smallest $\boldsymbol{R}$} & \textbf{Largest $\boldsymbol{R}$} \\
\midrule
dsj1000\_n999\_bounded-strongly-corr\_01 & 0.002836 & 0.003295 \\
dsj1000\_n999\_uncorr-similar-weights\_01 & 0.001736 & 0.001875 \\
dsj1000\_n999\_uncorr-similar-weights\_02 & 0.003298 & 0.003683 \\
dsj1000\_n999\_uncorr\_01 & 0.003325 & 0.003540 \\
\bottomrule
\end{tabularx}
\end{table}

All these values lie between 0.0017 and 0.0037. They are below 0.005, so they give 0.00 whether the generator rounded or
truncated. This agrees with the explanation of Section~\ref{sec:cause}, although it does not prove it, since the code
of the generator is not available. But no single one of them is the original value, and a statistic over several tours would only
be an estimate of it.

\subsection{The new renting rates}
\label{sec:new}
Instead of estimating the lost values, we define new ones. We keep rule~(\ref{eq:rule}) and fix every choice that it
leaves open, so that each new rent is fully specified and can be recomputed (Fig.~\ref{fig:flow}, bottom).
\begin{enumerate}
\item The packing is the optimal packing returned by the dynamic program of the 0-1 knapsack problem, run over the items
in the order of the file, where an item is packed only when it strictly increases the best profit. This rule breaks
ties between optimal packings in a fixed way, so the same packing is returned at every run. It does not mean that the
optimal packing is unique. We checked the optimal profit with Gurobi.
\item The tour is the tour of \texttt{linkern}, with its default settings and the random seed~1
(\texttt{linkern -Q -s 1 -o} \textit{output input}), started at the depot as in the code of the authors of the
benchmark. With our build of \texttt{linkern}, this command returns the same tour at every run, but another build may
not. The choice of the seed is arbitrary, and another seed would give another valid rent (Table~\ref{tab:spread}). For
these reasons, the published tour, not the command, defines the new rent.
\item The travel time is computed as in Section~\ref{sec:recover}, and the rent is the ratio of rule~(\ref{eq:rule}).
\end{enumerate}

Mathematically, the new rent is the optimal profit divided by the travel time of the reference solution, and this
reference solution has an objective of zero. The files store the double-precision value of this ratio in the shortest
decimal form that gives back the same double, not a value rounded to a fixed number of decimals. Evaluated with the stored rent, the
objective of the reference solution is zero up to floating-point rounding. We check it with an absolute tolerance of
$10^{-8}$, and the observed residuals are below $4\times10^{-11}$ on the four instances.

Table~\ref{tab:new} gives the result, with the travel time of the reference solution and the rent printed with six
significant digits. The tour of seed~1 has a length of 18,665,994 for the four instances, since they
share the same cities. The reference solution made of this tour and this packing restores, on each corrected instance,
the calibration intended by the original rule. The four reference solutions are published with the corrected files,
together with a script, \texttt{verify.py}, that checks them with these files only (see the Code
availability statement). It asserts that each reference
solution is feasible, that its packing is optimal, that the rent equals the optimal profit divided by its travel time,
and that its objective is zero within the tolerance. All four new rents are below 0.005, so they round to the value 0.00 of the original
files.

\begin{table}[ht]
\centering\footnotesize
\caption{New rents of the four instances.}
\label{tab:new}
\begin{tabularx}{\linewidth}{@{}Xrc@{}}
\toprule
\textbf{Instance} & \textbf{Travel time} & \textbf{New $\boldsymbol{R}$} \\
\midrule
dsj1000\_n999\_bounded-strongly-corr\_01 & 50,822,606 & 0.00284177 \\
dsj1000\_n999\_uncorr-similar-weights\_01 & 45,714,939 & 0.00187512 \\
dsj1000\_n999\_uncorr-similar-weights\_02 & 44,709,234 & 0.00368298 \\
dsj1000\_n999\_uncorr\_01 & 47,522,162 & 0.00348368 \\
\bottomrule
\end{tabularx}
\end{table}

These new rents are not the original ones, and results obtained with them cannot be compared with results obtained on the
original files. They give the four instances a positive rent of the size intended by the generator, computed by a rule
that is stated in full.

\subsection{Reference results on the corrected instances}
\label{sec:results}
To see how the four instances behave with and without a rent, we ran a published heuristic on the original file
($R=0.00$) and on the corrected file of each instance. We chose CoCo$^+$, the method of \citet{namazi2023solving} with the best
published values on the categorized instances. We used the public code of its authors, algorithm~4 (CoCoP). This code seeds its
random numbers with the clock, and we added an offset that depends on the run number, so that runs started at the same
time do not share their random numbers. It ran ten times on each file, with a budget of 600 seconds of wall-clock
time per run, which it used in full. Each run used one core, and eight runs
ran at a time on an Apple M1 Pro. Every solution was evaluated again on the \texttt{.ttp} file used by the run.
Table~\ref{tab:results} gives the results. On the corrected files, Min and Max are the worst and the best of the
ten runs, and the last column gives the smallest and largest number of picked items. Note that the value of CoCo$^+$ in Table~\ref{tab:published} is the best over
all runs of several CoCo variants, while the runs here use CoCoP alone.

\begin{table}[ht]
\centering\footnotesize
\caption{Results of CoCo$^+$ on the original and the corrected files.}
\label{tab:results}
\setlength{\tabcolsep}{4pt}
\begin{tabular}{@{}lrrrrrr@{}}
\toprule
& \textbf{Original file} & \multicolumn{5}{c}{\textbf{Corrected file}} \\
\cmidrule(l){3-7}
\textbf{Instance (dsj1000, 999 items)} & \textbf{Value} & \textbf{Min} & \textbf{Mean} & \textbf{Max} & \textbf{Std.\ dev.} & \textbf{Items picked} \\
\midrule
bounded strongly corr., 01 & 144,219 & 66,533 & 66,865 & 67,119 & 184 & 220 to 246 \\
similar weights, 01 & 85,719 & 33,100 & 33,463 & 34,145 & 322 & 90 \\
similar weights, 02 & 164,660 & 50,950 & 51,458 & 52,271 & 507 & 181 \\
uncorrelated, 01 & 165,494 & 57,036 & 57,187 & 57,535 & 165 & 233 to 234 \\
\bottomrule
\end{tabular}
\end{table}

On the original files, CoCo$^+$ behaves as on a knapsack problem. It returns the same value in all ten runs on each
instance, slightly below the optimal knapsack profit, at 144,219 on the categorized instance, the value
reported by five methods in Table~\ref{tab:published}. Since the rent is zero, the tour has no effect on the value.

On the corrected files, the picture changes. The ten runs give ten different values, with a standard deviation
between 165 and 507. The mean values are between 31\% and 46\% of the optimal knapsack profit, because the
travel time now has a cost. The packing also changes. On the bounded strongly correlated instance, CoCo$^+$ picks 220
to 246 items instead of the 291 of the optimal knapsack, and on the uncorrelated instance 233 or 234 instead of 257. With
fewer items, the thief carries less weight and travels faster. On the two instances with similar weights, every run
picked 90 and 181 items, as many as the optimal knapsack packings, although up to 91 and 182 items fit in the knapsack.
The profits of the packed items are lower than the optimal profit, so CoCo$^+$ picked other items than the optimal
knapsack, chosen together with the tour. The corrected instances behave as TTP instances again, in which the tour and
the packing must be chosen together. These values, with their minimum and maximum over ten runs, can be used by the
community as a reference for future comparisons on the corrected instances.

\section{Corrected files and the library \texttt{ttplib}}
\label{sec:library}
The original instances are text files in the \texttt{.ttp} format, which can be downloaded from the website of the
benchmark. To make them easier to use, we converted the 9,720 files to JSON, a common text format that most
programming languages can read directly. Each JSON file holds the header of the instance (name, type of knapsack,
number of cities and items, capacity, speeds, rent and distance type), the list of cities with their coordinates, and
the list of items with their profit, weight and city. For the 60 categorized instances and the three other instances
with $R=0.00$, we checked that the JSON files and the original files hold the same data in every field.

The corrected versions of the four instances are also JSON files. They are equal to the original instances except for
the rent, keep the original names, and are stored in a separate folder, so that the two versions cannot be mixed up. A
user who prefers the \texttt{.ttp} format can obtain a corrected file by copying the new rent of the JSON file into the
line \texttt{RENTING RATIO} of the original \texttt{.ttp} file.

All these files are stored in a public dataset on Hugging Face, a website that hosts data sets and makes them freely
available for download. The dataset has two folders: \texttt{instances}, with the 9,720 original instances, and
\texttt{corrected}, with the four corrected instances, their reference solutions and the scripts of
Section~\ref{sec:new}. A user can download any file from the dataset directly, without any software.

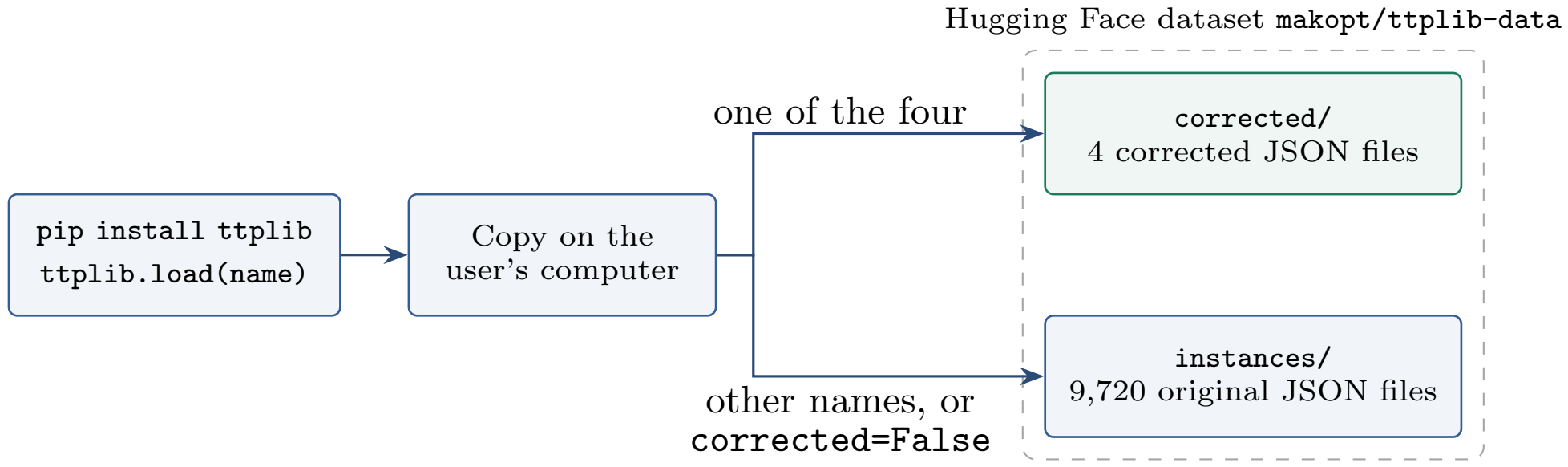
\begin{figure}[ht]
\centering
\begin{tikzpicture}[
  font=\footnotesize,
  box/.style={draw=libblue, fill=libblue!7, rounded corners=2pt, align=center, line width=0.5pt, minimum height=1.0cm},
  arr/.style={-{Stealth[length=2mm]}, line width=0.6pt, draw=libblue!80!black},
  lab/.style={font=\scriptsize, align=center, fill=white, inner sep=1pt}]
\node[box, text width=2.5cm] (user) at (0,0) {\texttt{pip install ttplib}\\[2pt]\texttt{ttplib.load(name)}};
\node[box, text width=2.3cm] (cache) at (3.2,0) {Copy on the\\user's computer};
\node[box, text width=3.2cm, draw=newgreen, fill=newgreen!7] (corr) at (8.9,1.0) {\texttt{corrected/}\\4 corrected JSON files};
\node[box, text width=3.2cm] (orig) at (8.9,-1.0) {\texttt{instances/}\\9,720 original JSON files};
\begin{scope}[on background layer]
\node[draw=black!35, dashed, rounded corners=3pt, fit=(corr)(orig), inner sep=5pt,
  label={[font=\footnotesize]above:Hugging Face dataset \texttt{makopt/ttplib-data}}] (hf) {};
\end{scope}
\draw[arr] (user) -- (cache);
\draw[arr] (cache.east) -- ++(3mm,0) |- (corr.west) node[lab, pos=0.65, above=1pt] {one of the four};
\draw[arr] (cache.east) -- ++(3mm,0) |- (orig.west) node[lab, pos=0.65, below=1pt] {other names, or\\\texttt{corrected=False}};
\end{tikzpicture}

\medskip
\begin{lrbox}{\codebox}
\begin{minipage}{0.92\linewidth}
\footnotesize
\begin{verbatim}
import ttplib                                   # after: pip install ttplib

names = ttplib.list_instances(category="all")   # the 60 categorized instances
inst = ttplib.load("dsj1000_n999_bounded-strongly-corr_01")
orig = ttplib.load("dsj1000_n999_bounded-strongly-corr_01", corrected=False)
print(inst.renting_ratio, orig.renting_ratio)   # 0.0028417669143102806 0.0
print(inst.capacity, inst.dimension, inst.num_items)    # 87826 1000 999
print(inst.get_distance(1, 2))                  # 709145
print(inst.get_items_at_node(2))     # [Item(index=1, profit=101.0, weight=1, ...)]
print(inst.calculate_tour_distance(range(1, inst.dimension + 1)))  # 557634042.0

for name, inst in ttplib.iter_instances(category="A"):   # the 20 instances of
    print(name, inst.renting_ratio)                     # category A, one at a time
\end{verbatim}
\end{minipage}
\end{lrbox}
\fcolorbox{libblue}{libblue!4}{\usebox{\codebox}}
\caption{How \texttt{ttplib} serves an instance (top), and an example of use with its output (bottom).}
\label{fig:ttplib}
\end{figure}

To make the instances even easier to use, we developed \texttt{ttplib}, a new open-source library for the Python
programming language (Fig.~\ref{fig:ttplib}). This note describes its version 0.2.2. Like most Python libraries, it is
published on PyPI, the public index of Python packages, and is installed with the single command
\texttt{pip install ttplib}. A program can then load any of the 9,720 instances by its name, for example with
\texttt{ttplib.load("dsj1000\_n999\_uncorr\_01")}. The library downloads the file of the instance from the dataset the
first time it is used and keeps a copy on the computer of the user, so that later uses need no connection. A user
therefore never has to download the whole benchmark. For the four instances with $R=0.00$, the library returns the
corrected version by default, and the original version with the option \texttt{corrected=False}.

The library also lists the instances, with filters on the TSP instance, the type of knapsack and the number of items.
For example, \texttt{ttplib.list\_instances(category="all")} returns the names of the 60 categorized instances, and
\texttt{category="A"}, \texttt{"B"} or \texttt{"C"} returns the 20 instances of one category. To work on many instances, \texttt{ttplib.iter\_instances()} loads them one
at a time, all 9,720 by default or those selected with the same filters, in their corrected or original version, and
frees the memory as it goes. For each instance, it
gives the distances between cities and the items of each city. It computes a distance only when it is needed, so that
even the largest instance, pla85900, loads in a few seconds with less than 200~MB of memory. The library comes with a
set of automatic tests, in the script \texttt{test\_ttplib.py}. They check the list of instances, the reading of an
instance, the distances and the items, the download from the dataset and the local copy, and the corrected versions of
the four instances.

Figure~\ref{fig:ttplib} shows how an instance is served and gives a short example of use. The README file of the
library, on its GitHub page, describes all its functions with their parameters, and the data and functions of an
instance. In the library, cities and items are numbered from 1, as in the files, so that city~1 is the depot. The links
to the dataset, to the library on PyPI and to its source code on GitHub are given in the Data availability and Code availability
statements.

\section{Conclusion and recommendations}
\label{sec:recommend}
Four of the 9,720 instances of the standard TTP benchmark have a renting rate of 0.00 in their files. On these
instances, the tour has no effect on the objective, and the problem reduces to a 0-1 knapsack problem. One of them is
among the 60 categorized instances used in most comparisons of TTP heuristics, where it has measured the quality of
knapsack heuristics rather than the coordination of tour and packing. The zeros are consistent with the two decimals
with which the rents are stored, and the same format may remove much of the precision of the rent of many other
instances, mainly those built on TSP instances with long tours. The original rents cannot be recovered from the
files and code published with the benchmark, so we defined new rents for the four instances with the rule of the original generator, applied
to a fully specified reference solution that is published with them. With these rents, the four instances behave as
TTP instances again. The corrected instances, the reference solutions and the scripts are freely available, and the
library \texttt{ttplib} gives access to all 9,720 instances, including the corrected ones, without downloading the
whole benchmark.

We make four recommendations for future work on this benchmark.
\begin{enumerate}
\item Studies that compare with published results should run the four instances with the original files ($R=0.00$),
since the published values were obtained on them, and say so. On the categorized instance, the optimal value
144,426 is then the best possible value. They should also run these instances with the corrected files and report
both results. The results on the corrected files will then build a base for comparisons on proper TTP instances in
future studies, starting from the reference results of Section~\ref{sec:results}.
\item Studies that do not compare with published values should use the corrected files for these four instances, and
state the version used.
\item Results on instances with a small rent should be read with care, since the stored rent of these instances may
have lost much of its precision (Section~\ref{sec:rounding}). Among the categorized instances, this concerns four instances with a
relative rounding half-width above 5\%.
\item Future generators of TTP instances should store the rent with enough significant digits to give back the computed
double-precision value, not with a fixed number of decimals. They should also publish the reference solution behind the
rent and a tolerance for checking it. A file can then be checked by evaluating this solution, whose objective must be
zero within the stated tolerance.
\end{enumerate}

\section*{Declarations}

\bmhead{Data availability} The original instances of the benchmark of \citet{polyakovskiy2014comprehensive} are
available from the website of the benchmark (\url{https://cs.adelaide.edu.au/~optlog/CEC2014COMP_InstancesNew/}). The
data generated in this study are openly available in the Hugging Face dataset \texttt{makopt/ttplib-data}
(\url{https://huggingface.co/datasets/makopt/ttplib-data}). The folder \texttt{instances} contains the 9,720 original
instances converted to JSON. The folder \texttt{corrected} contains the corrected versions of the four instances with a
zero renting rate, their reference solutions, the renting rates written in the 9,720 original files and the new renting
rates. The results of the heuristic runs summarized in Table~\ref{tab:results} are available from the corresponding
author on reasonable request.

\bmhead{Code availability} The Python library \texttt{ttplib}, version 0.2.2, is open source under the MIT license. It
is available from PyPI (\url{https://pypi.org/project/ttplib/}) and from GitHub
(\url{https://github.com/makopt/ttplib}), together with its documentation and its test script \texttt{test\_ttplib.py}.
The scripts used to compute the new renting rates (\texttt{recompute\_R.py}), to write the corrected files
(\texttt{write\_corrected.py}) and to verify them (\texttt{verify.py}) are available in the folder
\texttt{corrected/scripts} of the Hugging Face dataset. The runs of Section~\ref{sec:results} used the public code of
CoCo$^+$ provided by \citet{namazi2023solving}.

\bmhead{Author contribution} Mohammed Mizan Alanazi made the original finding of the zero renting rate. Mahdi
Khemakhem designed the study, wrote the code, carried out the experiments and wrote the manuscript. Both authors read
and approved the final manuscript.

\bibliography{note_refs}

\end{document}